\documentclass[conference]{IEEEtran}
\IEEEoverridecommandlockouts
\usepackage{cite}
\usepackage{amsmath,amssymb,amsfonts}
\usepackage{algorithmic}
\usepackage{graphicx}
\usepackage{textcomp}
\usepackage{xcolor}

\usepackage{multirow}
\usepackage{threeparttable}
\usepackage{bm}
\usepackage{svg}
\usepackage{amsmath}
\usepackage[normalem]{ulem}
\usepackage{cite}
\usepackage{makecell}
\usepackage[subrefformat=parens]{subcaption}

\def\BibTeX{{\rm B\kern-.05em{\sc i\kern-.025em b}\kern-.08em
    T\kern-.1667em\lower.7ex\hbox{E}\kern-.125emX}}
\begin{document}

\title{Imitation Learning Inputting Image Feature \\ to Each Layer of Neural Network}

\author{\IEEEauthorblockN{Koki Yamane}
\IEEEauthorblockA{\textit{Intelligent and Mechanical Interaction Systems} \\
\textit{University of Tsukuba}\\
Tsukuba, Japan \\
yamane.koki.td@alumni.tsukuba.ac.jp}
\and
\IEEEauthorblockN{Sho Sakaino}
\IEEEauthorblockA{\textit{Systems and Information Engineering} \\
\textit{University of Tsukuba}\\
Tsukuba, Japan \\
sakaino@iit.tsukuba.ac.jp}
\and
\IEEEauthorblockN{Toshiaki Tsuji}
\IEEEauthorblockA{\textit{Science and Engineering} \\
\textit{Saitama University}\\
Saitama, Japan \\
tsuji@ees.saitama-u.ac.jp}
}

\maketitle

\begin{abstract}
Imitation learning enables robots to learn and replicate human behavior from training data.
Recent advances in machine learning enable end-to-end learning approaches that directly process high-dimensional observation data, such as images.
However, these approaches face a critical challenge when processing data from multiple modalities, inadvertently ignoring data with a lower correlation to the desired output, especially when using short sampling periods.
This paper presents a useful method to address this challenge, which amplifies the influence of data with a relatively low correlation to the output by inputting the data into each neural network layer. 
The proposed approach effectively incorporates diverse data sources into the learning process.
Through experiments using a simple pick-and-place operation with raw images and joint information as input, significant improvements in success rates are demonstrated even when dealing with data from short sampling periods.
\end{abstract}


\section{Introduction}

Imitation learning~
\cite{rahmatizadeh2018vision},
\cite{10224318},
\cite{zitkovich2023rt},
\cite{zhao2023learning},
\cite{takeuchi2023motion}
is a method for learning behaviors from human-taught data.
In particular, end-to-end learning with high-dimensional observation data, such as direct image input, is attracting attention.
However, when data from multiple modalities are input simultaneously, data with a strong correlation to the output tend to be focused on, and data with a relatively weak correlation to the output tend to be ignored.
For example, if the joint angles of the robot at the current step and the images are input to predict the joint angles of the next step, the joint angles of the current step and those of the next step have a very strong correlation, and this characteristic is especially noticeable when the sampling period is short.
In such a case, even if the joint angle of the current step is used as the output as is, it is possible to achieve a rather small prediction error, and the learning results tend to be such that other inputs have little effect on the output.
To address this, we propose a method to increase the influence on the output by repeatedly inputting data with a relatively weak correlation to the output into each layer of the neural network.
We conducted experiments training a simple pick-and-place operation using raw images as input.
Experimental results confirmed that the success rate significantly improved with the proposed method, each layer input.
The proposed method allows us to effectively use data with weak correlation to the output, such as images, from data with short sampling periods.

The key contributions of this study are presented as follows.
\begin{itemize}
    \item We developed a simple neural network architecture for tasks that use two types of input with strong correlation and weak correlation to output, such as imitation learning using image and joint information.
    \item We analyze the difference between traditional and proposed architecture using gradient visualization.
\end{itemize}

The subsequent sections will delve into related work in the field, present our proposed method in detail, showcase experimental results, and summarize our contributions to imitation learning and neural network design.

\section{Related Works}

\begin{figure*}[!t]
    \centering
    \begin{minipage}[b]{0.49\linewidth}
        \centering
        \includegraphics[width=\linewidth]{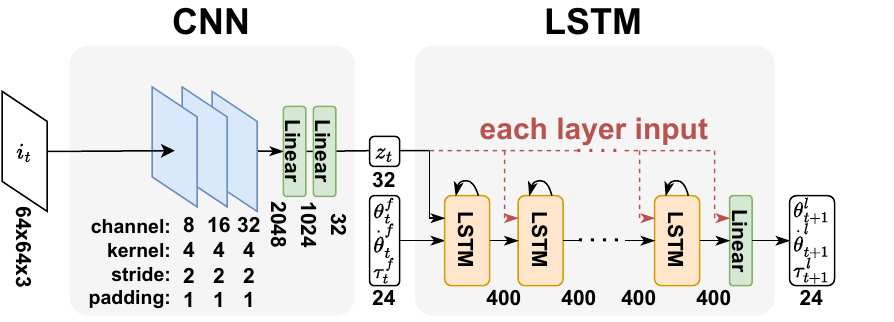}
        \subcaption{CNN + MLP}
    \end{minipage}
    \begin{minipage}[b]{0.49\linewidth}
        \centering
        \includegraphics[width=\linewidth]{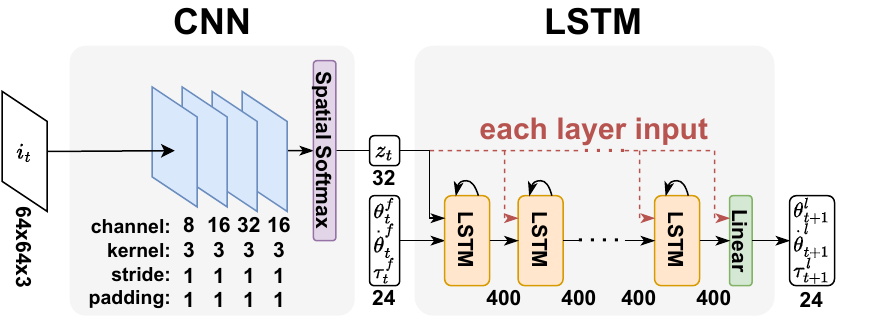}
        \subcaption{CNN + spatial softmax}
    \end{minipage}
    \caption{Neural network architecture}
    \label{fig:NN_models}
\end{figure*}

\subsection{Neural Network Architecture for Imitation Learning}
Recurrent neural network architectures, such as long short-term memory~(LSTM)~\cite{hochreiter1997long}, have long been used in robot imitation learning because they can remember past input and compute inference at high speed.
Although there have been some reports using image input~\cite{rahmatizadeh2018vision,10224318}, the operation frequency is low due to ignoring the image input described above and the limitation of the number of rememberable steps.
Although multilayering can increase the number of memorable steps, learning becomes difficult.
On the other hand, following the success of Transformer~\cite{vaswani2017attention} in natural language processing and image processing, there have been reports on the use of Transformer-based architectures in robot imitation learning~\cite{zitkovich2023rt,zhao2023learning}.
Although it is good at tasks with long-term dependencies and can use multimodal input such as image or language, inference computation is computationally expensive because the Transformer requires sequence input, including all previous step data at each step.
In addition, the Transformer may output noisy commands undesirable as control input for the robot because the self-attention layer used in the Transformer is not Lipschitz continuous~\cite{kim2021lipschitz}.
These problems become major issues when it is used at high frequency.
On the other hand, in this study, we develop neural network architecture using multilayer LSTM and each layer input.
The developed architecture can be used in high frequency with image input and is easy to learn.

\subsection{Each Layer Input}

In multilayer neural networks, inputting some inputs into multiple layers is occasionally used. For example, Wen \textit{et al.} used on natural language generation (NLG)~\cite{wen-etal-2015-semantically},
and Shridhar \textit{et al.} used to generate affordance map from RGB-D image and language command~\cite{shridhar2022cliport}.
In contrast, in this study, we focus on the fact that inputs with strong correlations with outputs and inputs with weak correlations are mixed in robot imitation learning and verify through experiments that inputting inputs with weak correlations to each layer can prevent them from being ignored.

\section{Method}

\subsection{Neural Network Architecture}

In this study, we used a network that combines a convolutional neural network~(CNN)~\cite{krizhevsky2012imagenet}, which compresses images into features, and a long short-term memory~(LSTM)~\cite{hochreiter1997long}, which inputs joint information and image features and outputs the next joint command value.
Experiments were conducted with two types of CNNs: a model with multi-layer perceptron (MLP) in the output layer (CNN~+~MLP) and a model with spatial softmax~\cite{finn2016deep} (CNN~+~spatial~softmax).
The entire picture of the neural network model is illustrated in Fig.~\ref{fig:NN_models}.
Where $i_t$ is the image input, $z_t$ is the image feature, $\theta$, $\dot{\theta}$, and $\tau$ denote the angle, angular velocity, and torque of each joint, and the subscripts $l$ and $f$ represent the leader and follower, respectively.

The CNN models had 64×64×3 image input and 32 output dimensions, rectified linear units~(ReLU)~\cite{nair2010rectified} as activation functions.
The CNN~+~MLP model had 3-layer convolutional layers with a 4×4 kernel and 2 strides compressing the image to half size, and 2-layer all connected layers with 1024 and 32 units.
This network could realize efficient computation by compressing image features and high representability by a huge parameter number of MLP.
However, because it directly connected CNN and MLP, this network lost the translation equivariance, an important feature of CNN.
The CNN~+~spatial~softmax model featured 4-layer convolutional layers using a 3×3 kernel and 1 stride, which did not compress the image.
In addition, spatial softmax was used as the output layer.
This network could realize useful features like the positions of objects with a small number of parameters and can use the translation equivariance of the CNN.
However, because it does not compress the image feature through a convolutional layer, the computation uses a large amount of memory and requires a longer training time.

The LSTM network had 56 input dimensions ([angle, angular velocity, torque] × [8 degrees of freedom] + [32 image features]), 6 layers of LSTM with 400 units, and a layer containing all connected layers with 24 units ([angle, angular velocity, torque] × [8 degrees of freedom]).
The inputs were the follower's response values for the current step, while the outputs were the follower's command values (leader's response value) for the next step.

\subsection{Each Layer Input}
Image features were input to each layer of LSTM and output layer by concatenating with the previous layer's output.
Each layer input has two important effects.

First, in the inference, it increases the effect of the input on the outputs.
Here, a multilayer neural network can be written as follows:
\begin{align}
&\boldsymbol{h}_n = f_n(\boldsymbol{h}_{n-1})\qquad(n = 1,\cdots,L) \label{eq:neural_network} \\
&\boldsymbol{x} = \boldsymbol{h}_0, \boldsymbol{y} = \boldsymbol{h}_L
\end{align}
where $\boldsymbol{h}_i$ and $f_i$ are an output and a function of the $i$-th layer of the neural network, $L$ is the number of layers of the neural network, $\boldsymbol{x}$ and $\boldsymbol{y}$ are an input and an output of the neural network, respectively.
Consider the partial differential coefficients as a measure of the influence of the inputs on the outputs.
The partial derivative of the output $\boldsymbol{y}$ with respect to the input $\boldsymbol{x}$ through a multilayer neural network is the product of the weights of each layer and the derivative of the activation function as follows:
\begin{equation}
\begin{aligned}
\frac{\partial \boldsymbol{y}}{\partial \boldsymbol{x}} = \frac{\partial \boldsymbol{h}_L}{\partial \boldsymbol{h}_0} &= \frac{\partial \boldsymbol{h}_L}{\partial \boldsymbol{h}_{L-1}}\frac{\partial \boldsymbol{h}_{L-1}}{\partial \boldsymbol{h}_{L-2}} \cdots \frac{\partial \boldsymbol{h}_1}{\partial \boldsymbol{h}_0} \\
&= \prod_{i=1}^L\frac{\partial \boldsymbol{h}_i}{\partial \boldsymbol{h}_{i-1}}.
\end{aligned}
\end{equation}
Therefore, adding more layers whose differential is less than 1, such as LSTM, reduces the partial differential coefficients.
On the other hand, a multilayer neural network using each layer input can be written as follows:
\begin{align}
&\boldsymbol{h}_n = f_n(\boldsymbol{h}_{n-1}, \boldsymbol{z})\qquad(n = 1,\cdots,L) \\
&\boldsymbol{x} \ne \boldsymbol{h}_0, \boldsymbol{y} = \boldsymbol{h}_L, \boldsymbol{z} = g(\boldsymbol{x})
\end{align}
where $g$ is a $C^1$-function.
In this case, the partial derivative of the output $\boldsymbol{y}$ with respect to the input $\boldsymbol{x}$ is the sum of the partial differentials up to each layer as follows:
\begin{align}
\frac{\partial \boldsymbol{h}_n}{\partial \boldsymbol{x}}
&= \frac{\partial f_n(\boldsymbol{h}_{n-1}, \boldsymbol{z})}{\partial \boldsymbol{x}} \notag \\
&= \frac{\partial f_n(\boldsymbol{h}_{n-1}, \boldsymbol{z})}{\partial \boldsymbol{h}_{n-1}} \frac{\partial \boldsymbol{h}_{n-1}}{\partial \boldsymbol{x}} + \frac{\partial f_L(\boldsymbol{h}_{n-1}, \boldsymbol{z})}{\partial \boldsymbol{z}}\frac{\partial \boldsymbol{z}}{\partial \boldsymbol{x}} \notag \\
&= \frac{\partial \boldsymbol{h}_n}{\partial \boldsymbol{h}_{n-1}}\frac{\partial \boldsymbol{h}_{n-1}}{\partial \boldsymbol{x}} + \frac{\partial \boldsymbol{h}_n}{\partial \boldsymbol{z}}\frac{\partial \boldsymbol{z}}{\partial \boldsymbol{x}} \\
\boldsymbol{\Delta}_n &:= \frac{\partial \boldsymbol{h}_n}{\partial \boldsymbol{x}}, \boldsymbol{A}_n := \frac{\partial \boldsymbol{h}_n}{\partial \boldsymbol{h}_{n-1}}, \boldsymbol{B}_n := \frac{\partial \boldsymbol{h}_n}{\partial \boldsymbol{z}}\frac{\partial \boldsymbol{z}}{\partial \boldsymbol{x}} \\
\boldsymbol{\Delta}_n &= \boldsymbol{A}_n\boldsymbol{\Delta}_{n-1} + \boldsymbol{B}_n \notag \\
&= \boldsymbol{A}_n(\boldsymbol{A}_{n-1}( \cdots (\boldsymbol{A}_1\boldsymbol{\Delta}_0 + \boldsymbol{B}_1) \cdots ) + \boldsymbol{B}_{n-1}) + \boldsymbol{B}_n \notag \\
&=\quad \boldsymbol{A}_n\boldsymbol{A}_{n-1}\cdots\boldsymbol{A}_3\boldsymbol{A}_2\boldsymbol{B}_1 \notag \\
&\quad + \boldsymbol{A}_n\boldsymbol{A}_{n-1}\cdots\boldsymbol{A}_3\boldsymbol{B}_2 \notag \\
&\hspace{2.5em} \vdots \notag \\
&\quad + \boldsymbol{A}_n\boldsymbol{B}_{n-1} \notag \\
&\quad + \boldsymbol{B}_n \hspace{30mm} \left( \because \boldsymbol{\Delta}_0=\frac{\partial \boldsymbol{h}_0}{\partial \boldsymbol{x}}=\boldsymbol{O} \right) \notag \\
&= \sum_{j=1}^n \biggl( \prod_{i=j+1}^n \boldsymbol{A}_i \biggr) \boldsymbol{B}_j \\
\therefore \frac{\partial \boldsymbol{y}}{\partial \boldsymbol{x}} &= \frac{\partial \boldsymbol{h}_L}{\partial \boldsymbol{x}} = \boldsymbol{\Delta}_L = \sum_{j=1}^L \biggl( \prod_{i=j+1}^L\frac{\partial \boldsymbol{h}_i}{\partial \boldsymbol{h}_{i-1}} \biggr) \frac{\partial \boldsymbol{h}_j}{\partial \boldsymbol{z}}\frac{\partial \boldsymbol{z}}{\partial \boldsymbol{x}}. \label{eq:grad_each_layer_input}
\end{align}
It allows the network to have large partial differential coefficients even when the entire network is multilayered.

Second, in the training, it avoids gradient vanishment and is easy to learn by backpropagation.
This phenomenon is caused by the fact that the same principle of increasing the influence of the input, as described above, also applies when using the error backpropagation method.
Here, the partial derivative of loss function $E(\boldsymbol{y})$ with respect to the weights of the $k$-th layer $\boldsymbol{W}_k$ of typical neural network is as follows:
\begin{equation}
\begin{aligned}
\frac{\partial E(\boldsymbol{y})}{\partial \boldsymbol{W}_k} &= \frac{\partial E(\boldsymbol{y})}{\partial \boldsymbol{y}}\frac{\partial \boldsymbol{y}}{\partial \boldsymbol{h}_k}\frac{\partial \boldsymbol{h}_k}{\partial \boldsymbol{W}_k} \\
&= \frac{\partial E(\boldsymbol{y})}{\partial \boldsymbol{y}} \biggr( \prod_{i=k+1}^L\frac{\partial \boldsymbol{h}_i}{\partial \boldsymbol{h}_{i-1}} \biggl) \frac{\partial \boldsymbol{h}_k}{\partial \boldsymbol{W}_k}.
\end{aligned}
\end{equation}
Since the gradient vanishment is pronounced in a multilayer LSTM, a CNN coupled to the input layer will have very small gradients when trained using error backpropagation, making learning difficult.
In contrast, in the same way as Eq.~(\ref{eq:grad_each_layer_input}), the partial derivative of the loss function $E(\boldsymbol{y})$ with respect to the weights of the $k$-th layer $\boldsymbol{W}_k$ whose output $\boldsymbol{x}=\boldsymbol{h}_k$ is input to each layer is as follows:
\begin{equation}
\begin{aligned}
\frac{\partial E(\boldsymbol{y})}{\partial \boldsymbol{W}_k} &= \frac{\partial E(\boldsymbol{y})}{\partial \boldsymbol{y}}\frac{\partial \boldsymbol{y}}{\partial \boldsymbol{x}}\frac{\partial \boldsymbol{x}}{\partial \boldsymbol{W}_k} \\
&=\frac{\partial E(\boldsymbol{y})}{\partial \boldsymbol{y}} \Biggl\{ \sum_{j=k+1}^L \biggl( \prod_{i=j+1}^L\frac{\partial \boldsymbol{h}_i}{\partial \boldsymbol{h}_{i-1}} \biggr) \frac{\partial \boldsymbol{h}_j}{\partial \boldsymbol{z}}\frac{\partial \boldsymbol{z}}{\partial \boldsymbol{x}} \Biggr\} \frac{\partial \boldsymbol{x}}{\partial \boldsymbol{W}_k}.
\end{aligned}
\end{equation}
Therefore, when the output of the CNN is input to each layer of the LSTM, the learning process is expected to be smooth because large gradients are propagated from the layer close to the output layer.
Although this phenomenon is similar to the learning of a multilayer neural network by residual learning~\cite{he2016deep}, the main difference is that each layer input is applied to the specific input whose influence wants to be increased instead of to all elements.

\section{Experimental setup}

\begin{figure}[!t]
        \begin{minipage}[b]{0.425\linewidth}
            \centering
            \includegraphics[width=\linewidth]{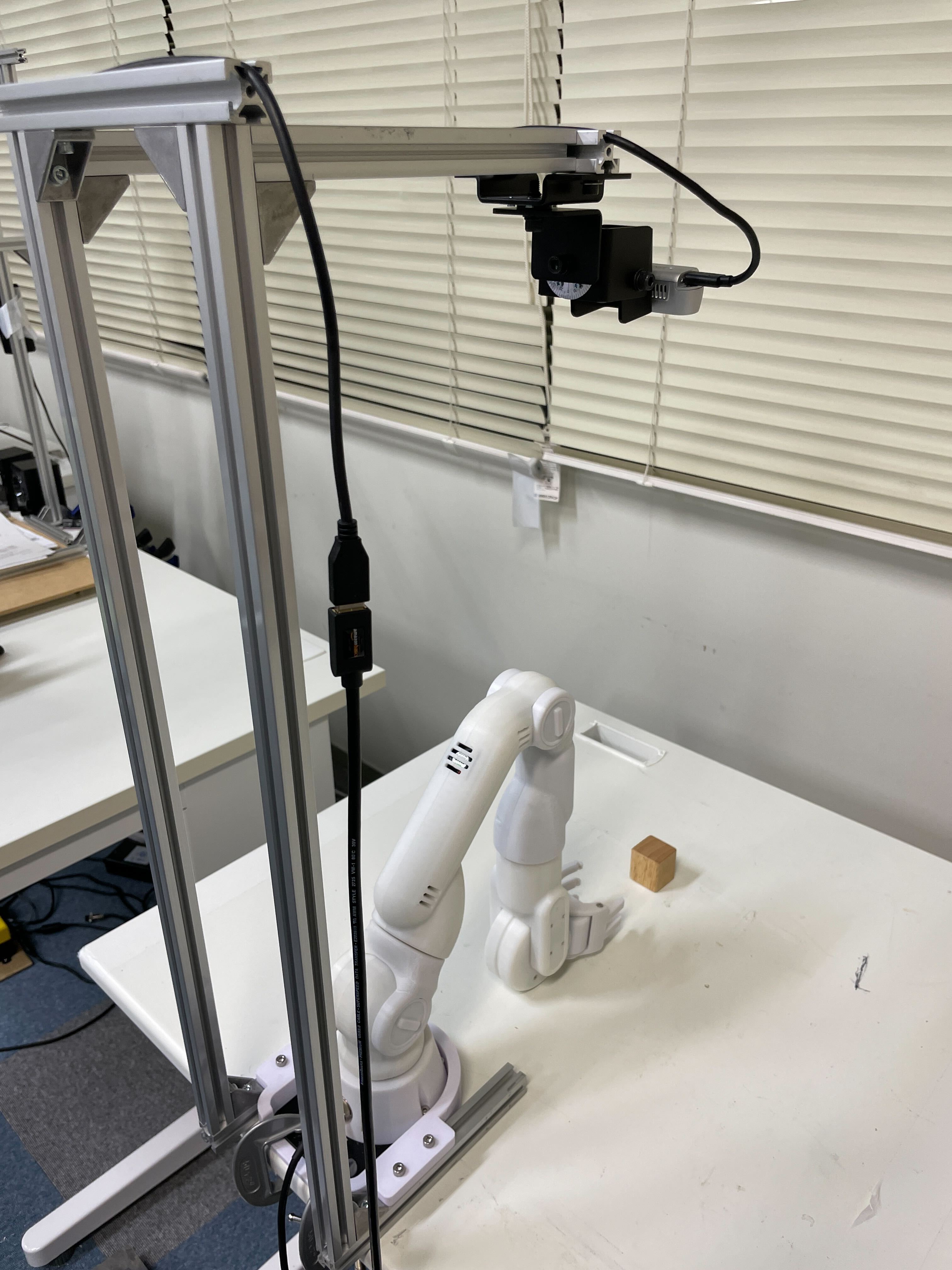}
            \subcaption{Hardware setup}
            \label{fig:environment}
        \end{minipage}
        \begin{minipage}[b]{0.565\linewidth}
            \centering
            \includegraphics[width=\linewidth]{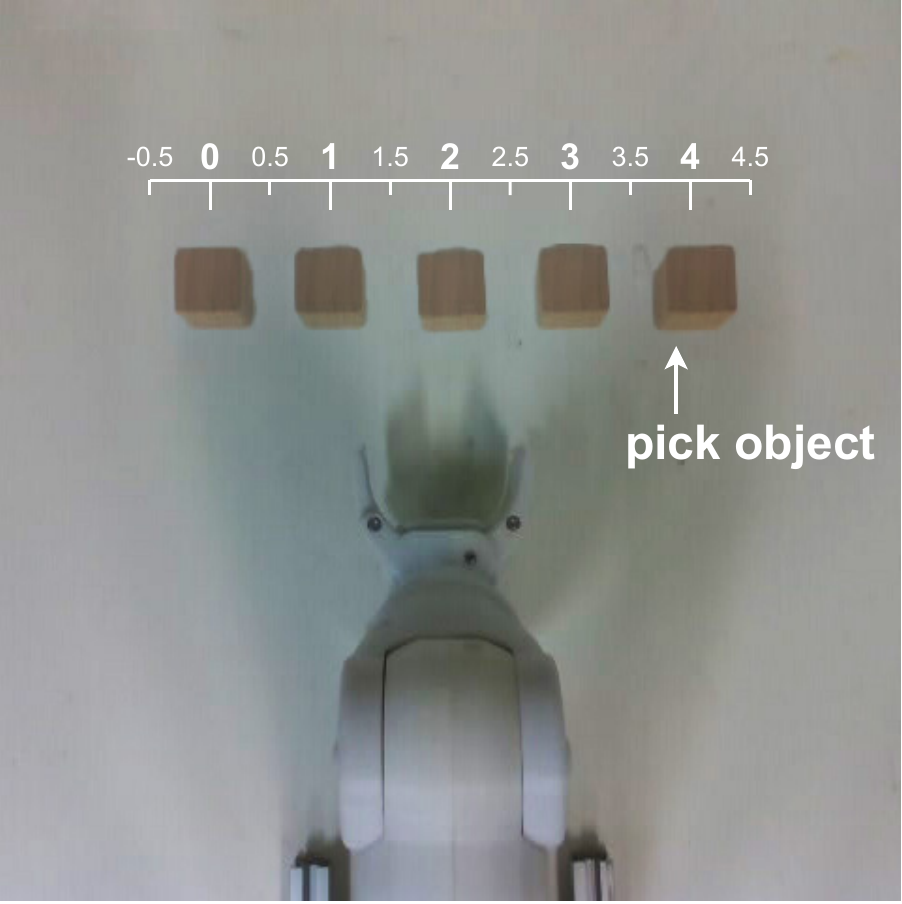}
            \subcaption{Object's initial position}
            \label{fig:object_initial_position}
        \end{minipage}
        \caption{Experimantal setups}
\end{figure}

\subsection{Robot System}
CRANE-X7, a manipulator manufactured by RT Corporation, was employed.
The manipulator exhibits seven degrees of freedom, while the gripper exhibits one degree of freedom, thereby providing a total of eight degrees of freedom.
We replaced the robot's hand with a cross-structure hand~\cite{10325570} without finger holes. 
We also used an Intel RealSense D435i to capture RGB images.
The robot and camera were connected to a PC running Ubuntu 22.04 LTS.
The PC was equipped with 32 GB RAM, an AMD Ryzen 7 3700X 8-Core Processor as CPU, and a GeForce RTX 2080 SUPER (rev a1) as GPU.
The overview of the hardware setup is presented in Fig.~\ref{fig:environment}.
Each axis of the manipulator was controlled using a position and force hybrid controller that was used in the related works~\cite{saigusa2022imitation,10325570}.
To collect demonstration data for imitation learning, we employed 4-channel bilateral control~\cite{sakaino2011multi,shikata2023modal}, which is a system that comprises two robots: a leader directly manipulated by a human and a follower of the leader.
The system controls the position and torque of the two robots to synchronize them.
Since the 7 DoF manipulator is difficult to control by human operation, joint 2 was fixed by position control and the robot was used as a 6 DoF manipulator.

\begin{table*}[!t]
    \caption{Success rate}
    \label{table:success_rate}
    \centering
    \begin{tabular}{c|c|ccccccccccc|c}
    \hline
    model & each layer input & -0.5 & 0 & 0.5 & 1 & 1.5 & 2 & 2.5 & 3 & 3.5 & 4 & 4.5 & total \\
    \hline
    \hline
    \multirow{2}{*}{CNN + MLP} & w/o & 20 & 80 & $\mathbf{100}$ & 80 & 20 & 0 & 0 & 0 & 0 & 0 & 0 & 29.1 \\
    & w/ & $\mathbf{100}$ & 60 & $\mathbf{100}$ & $\mathbf{100}$ & $\mathbf{100}$ & 80 & $\mathbf{100}$ & $\mathbf{100}$ & 80 & 60 & 0 & $\mathbf{80.0}$ \\
    \hline
    \multirow{2}{*}{CNN + spatial softmax} & w/o & 80 & 40 & 60 & 0 & 60 & 0 & 20 & 0 & 0 & 0 & 0 & 23.6 \\
    & w/ & $\mathbf{100}$ & $\mathbf{100}$ & $\mathbf{100}$ & $\mathbf{100}$ & $\mathbf{100}$ & 80 & $\mathbf{100}$ & $\mathbf{100}$ & 80 & 80 & $\mathbf{100}$ & $\mathbf{94.5}$ \\
    \hline
    \end{tabular}
\end{table*}

\begin{figure*}[!t]
    \centering
    \begin{minipage}[b]{0.49\linewidth}
        \centering
        \includegraphics[width=\linewidth]{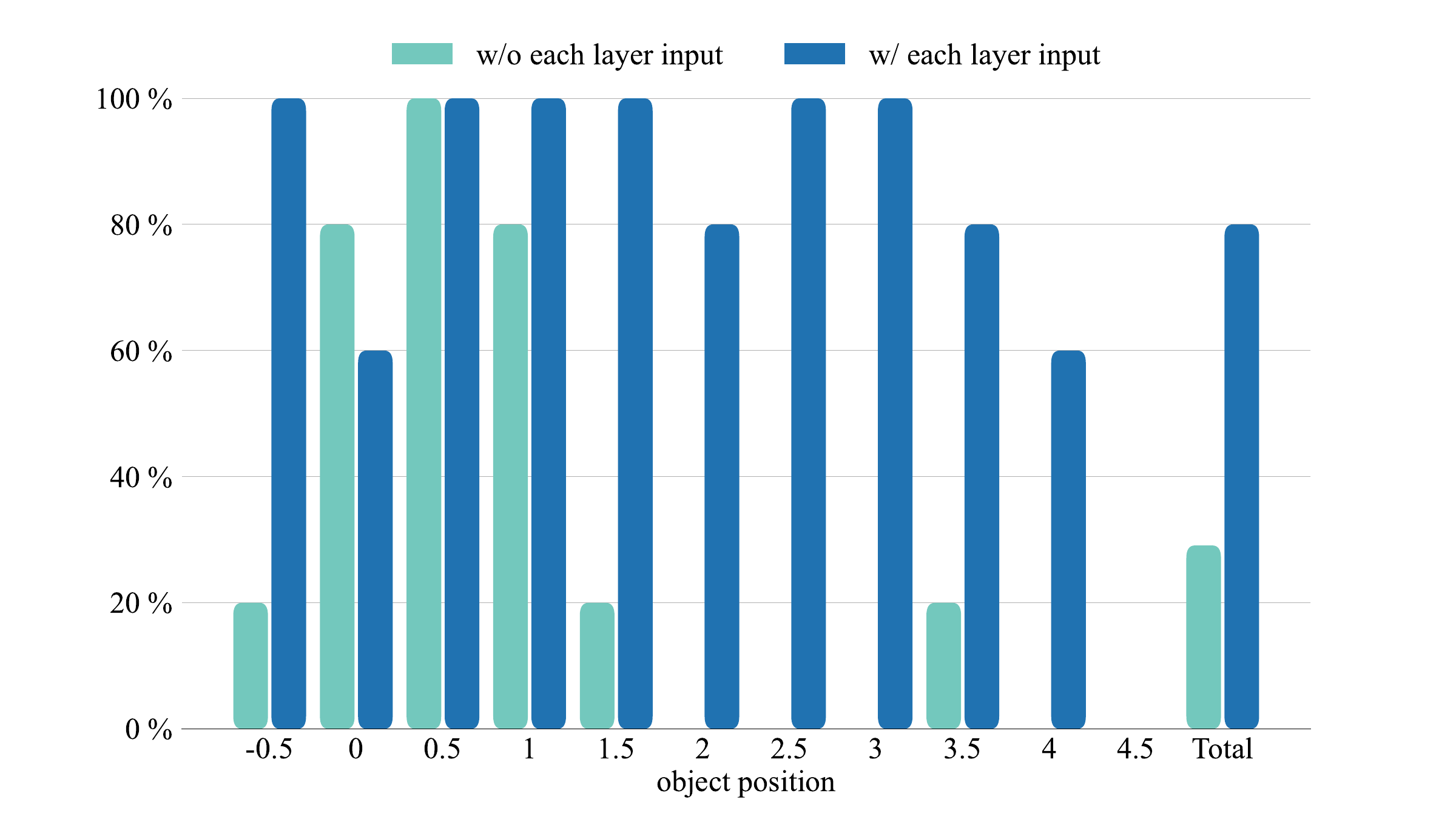}
        \subcaption{CNN + MLP}
    \end{minipage}
    \begin{minipage}[b]{0.49\linewidth}
        \centering
        \includegraphics[width=\linewidth]{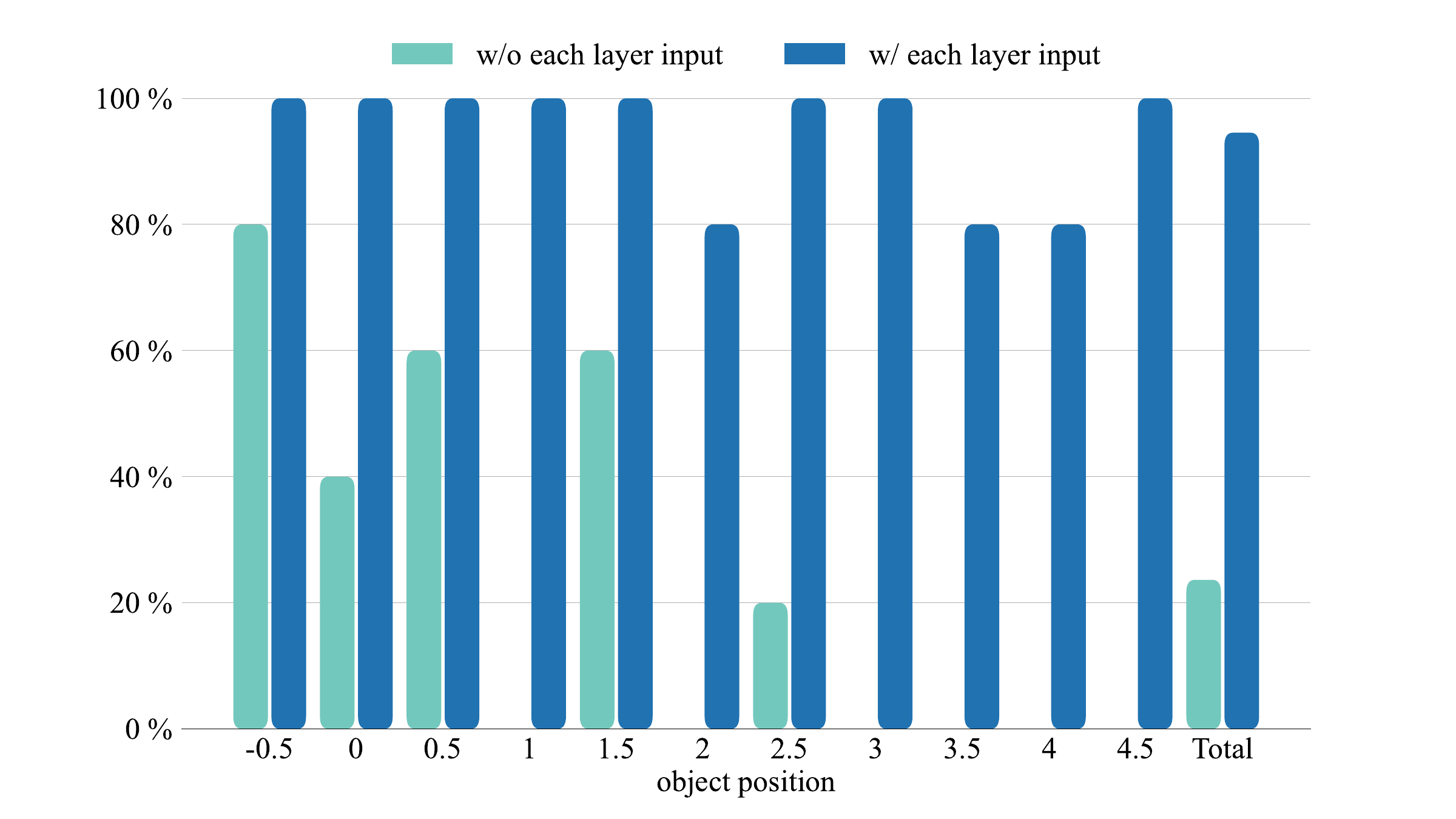}
        \subcaption{CNN + spatial softmax}
    \end{minipage}
    \caption{Success rate}
    \label{fig:success_rate}
\end{figure*}

\subsection{Task Design}

The robot picked up the object and placed it in the center of the workspace.
Success was determined by whether the object floated off the table.
For training, we used 5 initial positions of the object, and for evaluation, we used 11 initial positions, including each intermediate position of the trained position.
The evaluation trials were performed 5 times for each setup.
The initial positions of the objects used to collect the demonstration data and the evaluation test are shown in Fig.~\ref{fig:object_initial_position}.

\subsection{Dataset}
Demonstration data were collected using 5 initial positions of the object, and we collected 4 training datasets and 1 validation dataset per object (in total, 20 training data and 5 validation data).
Images were acquired at 50 Hz and joint information was acquired at 500 Hz.
The joint information acquired at 500 Hz was downsampled to 50 Hz via sampling every 10 steps.
When training a neural network with time-series data, the learning efficiency can be improved by reducing the sampling frequency to a certain degree~\cite{rahmatizadeh2018vision}.
In addition, the number of data can be increased by a factor of 10 (therefore, we used 200 training data and 50 validation data).
Furthermore, in combination with a low-pass filter, high-frequency components that do not require imitation can be eliminated from the data, thereby increasing the stability of the operation.
The length of each sequence was aligned to the same length by padding, which copies the last value.
The length after padding was set to the mean of the sequence lengths plus twice the standard deviation.

\subsection{Training}
Mean Squared Error (MSE) was used for the loss function, and Adam~\cite{kingma2014adam} was used for optimization.
The training rate was 0.001.
5000 epochs were trained with a batch size of 16.
The data were normalized before input to the neural network by setting the mean to zero, and the standard deviation to one, and the output of the neural network was denormalized before computing the loss function, except for fixed joint 2.
For data augmentation, normally distributed noise with a variance of 0.01 was added to the input for each training step before normalization.
The training was computed on an AMD EPYC 7763 64-Core Processor and NVIDIA RTX A6000.
The training took approximately 10 hours for the CNN~+~MLP model and 24 hours for the CNN~+~spatial~softmax model.

\begin{figure*}[!t]
    \centering
    \includegraphics[width=\linewidth]{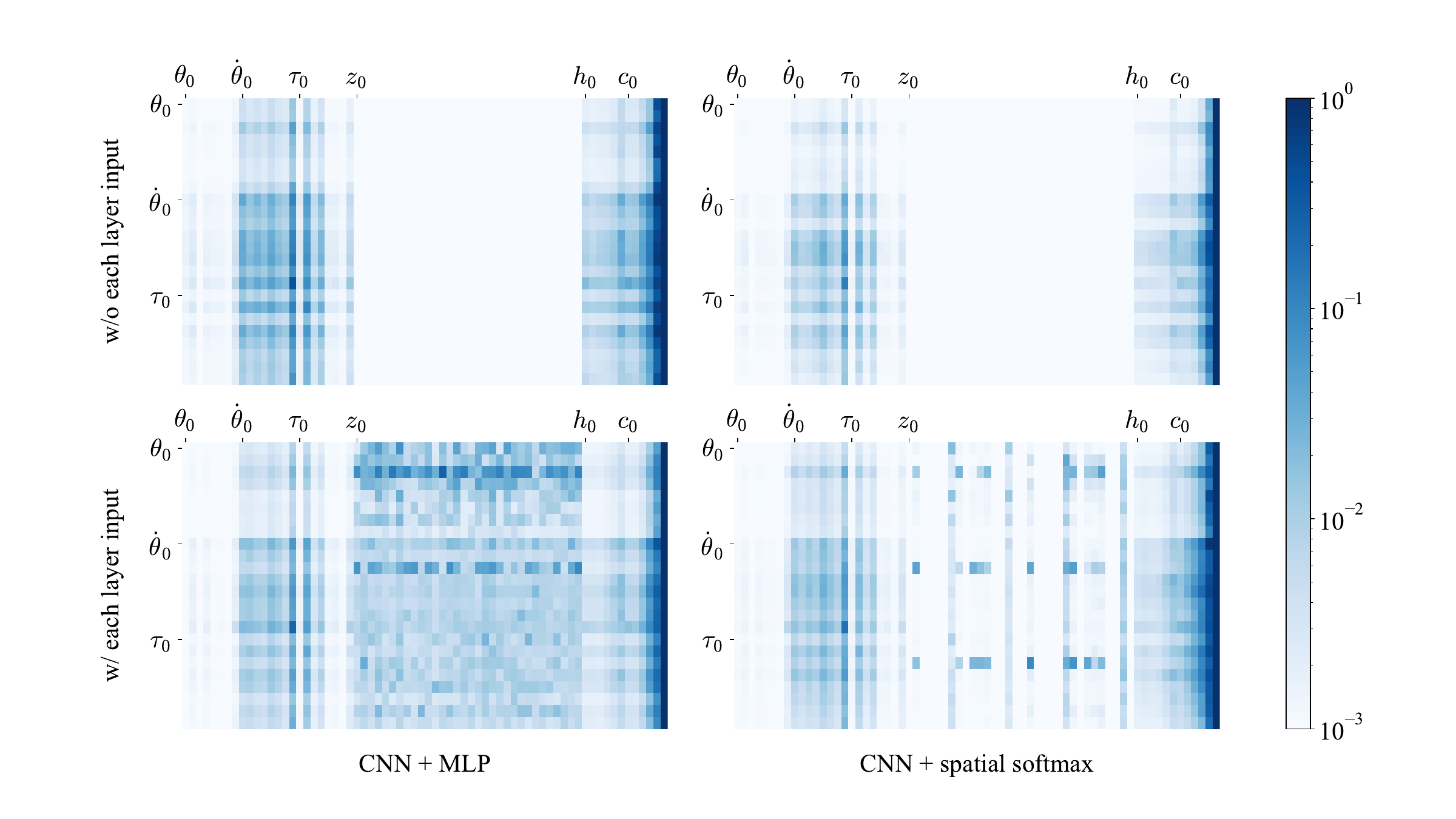}
    \caption{Attribution matrix of LSTM}
    \label{fig:attribution_matrix}
\end{figure*}

\section{Experiment results}
\subsection{Success rate}
The success rates are presented in Table~\ref{table:success_rate} and Fig.~\ref{fig:success_rate}.
Both models achieved significantly higher success rates with each layer input than without.
In the case of the model with each layer input, CNN~+~MLP had a 50.9\% higher average success rate, and CNN~+~spatial~softmax had a 70.9\% higher average success rate than the case of the model without each layer input.
In the case of the model without each layer input, although the robot achieved a high success rate when the object was placed on the left side, the initial position on the right side was rarely successful.
The robot moved toward the left side even when the object was placed on the right side, suggesting that the model without each layer input ignores the image input and moves in the same direction each time, regardless of the object's initial position.
In contrast, the model with each input layer achieved a high success rate for all initial positions, confirming that the robot could identify the object's position from the image input and output movements accordingly.

In the case of the model with each layer input, the CNN~+~spatial~softmax model had an average success rate 14.5\% higher than the CNN~+~MLP model.
In addition, the CNN~+~MLP model with input from each layer showed a success rate of 0\% for the initial object position 4.5 and no extrapolation, while the CNN~+~spatial~softmax model with input from each layer showed a success rate of 100\% for the initial object position 4.5 and a high extrapolation performance.
This may be because the CNN~+~spatial~softmax model has the property of parallel shift assimilation, i.e., the image features are also shifted when the input image is shifted in parallel.

\subsection{Attribution matrix}
In general, if the relation between the $n$-dimensional vector $\boldsymbol{x}=[x_0, x_1, \cdots, x_n]^T \in \mathbb{R}^n$ and the $m$-dimensional vector $\boldsymbol{y}=[y_0, y_1, \cdots, y_m]^T \in \mathbb{R}^m$ is $C^1$-function like
\begin{align}
y_j=f_j(x_0, x_1, \cdots, x_n)\qquad(j = 0,1,\cdots,m),
\end{align}
output differentiation~$\boldsymbol{\dot{y}}$ is expressed by Jacobean matrix~$\bm{J}_{y}(\boldsymbol{x})=\frac{\partial \boldsymbol{y}}{\partial \boldsymbol{x}}$ as follows:
\begin{align}
\boldsymbol{\dot{y}}&=\bm{J}_{y}(\boldsymbol{x})\boldsymbol{\dot{x}} \\
\label{eq:jacobean}
\begin{bmatrix}
\frac{d y_0}{dt} \\
\frac{d y_1}{dt} \\
\vdots \\
\frac{d y_m}{dt} \\
\end{bmatrix}
&= 
\begin{bmatrix}
\frac{\partial y_0}{\partial x_0} & \frac{\partial y_0}{\partial x_1} & \cdots & \frac{\partial y_0}{\partial x_n} \\
\frac{\partial y_1}{\partial x_0} & \frac{\partial y_1}{\partial x_1} & \cdots & \frac{\partial y_1}{\partial x_n} \\
\vdots & \vdots & \ddots & \vdots \\
\frac{\partial y_m}{\partial x_0} & \frac{\partial y_m}{\partial x_1} & \cdots & \frac{\partial y_m}{\partial x_n} \\
\end{bmatrix}
\begin{bmatrix}
\frac{d x_0}{dt} \\
\frac{d x_1}{dt} \\
\vdots \\
\frac{d x_n}{dt} \\
\end{bmatrix} \notag \\
&=
\begin{bmatrix}
\frac{\partial y_0}{\partial x_0} \frac{d x_0}{dt} + \frac{\partial y_0}{\partial x_1} \frac{d x_1}{dt} + \cdots + \frac{\partial y_0}{\partial x_n} \frac{d x_n}{dt} \\
\frac{\partial y_1}{\partial x_0} \frac{d x_0}{dt} + \frac{\partial y_1}{\partial x_1} \frac{d x_1}{dt} + \cdots + \frac{\partial y_1}{\partial x_n} \frac{d x_n}{dt} \\
\vdots \\
\frac{\partial y_m}{\partial x_0} \frac{d x_0}{dt} + \frac{\partial y_m}{\partial x_1} \frac{d x_1}{dt} + \cdots + \frac{\partial y_m}{\partial x_n} \frac{d x_n}{dt} \\
\end{bmatrix}.
\end{align}
Equation~(\ref{eq:jacobean}) can be interpreted as additive decomposition of the output differentiation, and each term represents the effect of each input to output.
Here, we define the attribution matrix~$\mathbf{A}$ using the absolute value of each term as follows:
\begin{align}
\mathbf{A} = 
\begin{bmatrix}
|\frac{\partial y_0}{\partial x_0} \frac{d x_0}{dt}| & |\frac{\partial y_0}{\partial x_1} \frac{d x_1}{dt}| & \cdots & |\frac{\partial y_0}{\partial x_n} \frac{d x_n}{dt}| \\
|\frac{\partial y_1}{\partial x_0} \frac{d x_0}{dt}| & |\frac{\partial y_1}{\partial x_1} \frac{d x_1}{dt}| & \cdots & |\frac{\partial y_1}{\partial x_n} \frac{d x_n}{dt}| \\
\vdots & \vdots & \ddots & \vdots \\
|\frac{\partial y_m}{\partial x_0} \frac{d x_0}{dt}| & |\frac{\partial y_m}{\partial x_1} \frac{d x_1}{dt}| & \cdots & |\frac{\partial y_m}{\partial x_n} \frac{d x_n}{dt}| \\
\end{bmatrix}.
\end{align}

We have visualized the attribution matrix of the LSTM of each model in Fig.~\ref{fig:attribution_matrix}.
Where $\theta$, $\dot{\theta}$, and $\tau$ denote the angle, angular velocity, and torque of each joint, $z$ is the image feature, $h$ and $c$ are the output and memory cell of each LSTM layer, respectively.
The visualized results are the average of all evaluation trials and all time steps.
The time derivative was approximated by the backward derivative.
To visualize the relative influence of each input, it was normalized by dividing by the maximum value of each row.
Since the LSTM cell and hidden have many elements, only the maximum value of each layer was visualized.

In the case of the model without each layer input, the influence of all dimensions of the image features on the output was very small.
In contrast, in the model with each layer input, the image features greatly influenced the output.
The differences between the two network architectures show that in the CNN~+~MLP model, all 32 dimensions of the image features had a large influence, while in the CNN~+~spatial~softmax model, only a few dimensions had a large influence, while the other dimensions were ignored.
This suggests that spatial softmax could efficiently capture features represented in small image dimensions, and this difference may have caused the CNN~+~spatial~softmax model to have a higher task success rate than the CNN~+~MLP model.

The spatial softmax features are obtained in the form of $xy$ coordinates, and in this visualization, they are arranged in the order $([z_0, z_1, ..., z_{31}] = [x_0, x_1, ..., x_{15}, y_0, y_1, ..., y_{15}])$.
The attribution matrix shows a similar pattern in the $x$ and $y$ coordinate regions, indicating that the gradient increases in the corresponding dimension.
Interestingly, the feature corresponding to the $y$-coordinate had a large influence, even though the object's initial position was only changed on the $x$-axis in this task.
This suggests that the robot looks at the position of the object along the path as well as the initial position of the object to check how far the task is progressing.

Looking at the input of joint information, it was found that angular velocity was a particularly large contributor, while angle was not used much.
The investigation suggests that the angular velocity was integrated internally in LSTM and had the angle estimated value.

Overall, the input that contributed the most to the output was the memory cell of the LSTM ($c$ in the figure).
In this experiment, the training data was not clipped in time direction, and the task was trained to start when the internal state of the LSTM was $\mathbf{0}$. 
In addition, the LSTM was used with a large number of parameters.
It is possible that the output was based on time, e.g., the output until the end of the task was embedded in the weights of the network, and the time was counted within the LSTM to determine the current output.
If this were the case, it would be difficult to change the behavior flexibly according to the input during the task.
Although the current task was a problem setting where only the initial placement of objects changed, verification in tasks where the situation changes during task execution is a future issue.
In addition, it needs to be verified whether this tendency changes when the training data is randomly clipped in the temporal direction.

\section{Conclusions}

In this study, we proposed a method to increase the influence of elements with weak correlations with outputs by inputting them into each layer of a neural network to solve the problem that elements with weak correlations with outputs are sometimes ignored in imitation learning using inputs from multiple modalities.
Experiments on a simple pick-and-place imitation learning task, in which images and joint information are used as inputs, and the next joint information is used as outputs, confirmed that the proposed method significantly increases the success rate of the task.
In addition, by visualizing the gradient of each input/output of the neural network, it was confirmed that the gradient of the element that was input in each layer became significantly larger before and after the introduction of the proposed method, and the effect on the output became larger.
As a future issue, although all layers were input in this case, it is necessary to verify whether the same effect can be obtained when inputs are added not to all layers but to some layers or only to the final layer.

\section*{ACKNOWLEDGMENT}
This study is supported by the Japan Society for the Promotion of Science through a Grant-in-Aid for Scientific Research (B) under Grant 21H01347.

\bibliographystyle{ieeetr}
\bibliography{reference.bib}

\begin{thebibliography}{10}

\bibitem{rahmatizadeh2018vision}
R.~Rahmatizadeh, P.~Abolghasemi, L.~B{\"o}l{\"o}ni, and S.~Levine, ``Vision-based multi-task manipulation for inexpensive robots using end-to-end learning from demonstration,'' in {\em 2018 IEEE international conference on robotics and automation (ICRA)}, pp.~3758--3765, IEEE, 2018.

\bibitem{10224318}
T.~Hara, T.~Sato, T.~Ogata, and H.~Awano, ``Uncertainty-aware haptic shared control with humanoid robots for flexible object manipulation,'' {\em IEEE Robotics and Automation Letters}, vol.~8, no.~10, pp.~6435--6442, 2023.

\bibitem{zitkovich2023rt}
B.~Zitkovich, T.~Yu, S.~Xu, P.~Xu, T.~Xiao, F.~Xia, J.~Wu, P.~Wohlhart, S.~Welker, A.~Wahid, {\em et~al.}, ``Rt-2: Vision-language-action models transfer web knowledge to robotic control,'' in {\em Conference on Robot Learning}, pp.~2165--2183, PMLR, 2023.

\bibitem{zhao2023learning}
T.~Z. Zhao, V.~Kumar, S.~Levine, and C.~Finn, ``Learning fine-grained bimanual manipulation with low-cost hardware,'' {\em Robotics: Science and Systems}, 2023.

\bibitem{takeuchi2023motion}
K.~Takeuchi, S.~Sakaino, and T.~Tsuji, ``Motion generation based on contact state estimation using two-stage clustering,'' {\em IEEJ Journal of Industry Applications}, p.~22012635, 2023.

\bibitem{hochreiter1997long}
S.~Hochreiter and J.~Schmidhuber, ``Long short-term memory,'' {\em Neural computation}, vol.~9, no.~8, pp.~1735--1780, 1997.

\bibitem{vaswani2017attention}
A.~Vaswani, N.~Shazeer, N.~Parmar, J.~Uszkoreit, L.~Jones, A.~N. Gomez, {\L}.~Kaiser, and I.~Polosukhin, ``Attention is all you need,'' {\em Advances in neural information processing systems}, vol.~30, 2017.

\bibitem{kim2021lipschitz}
H.~Kim, G.~Papamakarios, and A.~Mnih, ``The lipschitz constant of self-attention,'' in {\em International Conference on Machine Learning}, pp.~5562--5571, PMLR, 2021.

\bibitem{wen-etal-2015-semantically}
T.-H. Wen, M.~Ga{\v{s}}i{\'c}, N.~Mrk{\v{s}}i{\'c}, P.-H. Su, D.~Vandyke, and S.~Young, ``Semantically conditioned {LSTM}-based natural language generation for spoken dialogue systems,'' in {\em Proceedings of the 2015 Conference on Empirical Methods in Natural Language Processing} (L.~M{\`a}rquez, C.~Callison-Burch, and J.~Su, eds.), (Lisbon, Portugal), pp.~1711--1721, Association for Computational Linguistics, Sept. 2015.

\bibitem{shridhar2022cliport}
M.~Shridhar, L.~Manuelli, and D.~Fox, ``Cliport: What and where pathways for robotic manipulation,'' in {\em Conference on Robot Learning}, pp.~894--906, PMLR, 2022.

\bibitem{krizhevsky2012imagenet}
A.~Krizhevsky, I.~Sutskever, and G.~E. Hinton, ``Imagenet classification with deep convolutional neural networks,'' {\em Advances in neural information processing systems}, vol.~25, 2012.

\bibitem{finn2016deep}
C.~Finn, X.~Y. Tan, Y.~Duan, T.~Darrell, S.~Levine, and P.~Abbeel, ``Deep spatial autoencoders for visuomotor learning,'' in {\em 2016 IEEE International Conference on Robotics and Automation (ICRA)}, pp.~512--519, IEEE, 2016.

\bibitem{nair2010rectified}
V.~Nair and G.~E. Hinton, ``Rectified linear units improve restricted boltzmann machines,'' in {\em Proceedings of the 27th international conference on machine learning (ICML-10)}, pp.~807--814, 2010.

\bibitem{he2016deep}
K.~He, X.~Zhang, S.~Ren, and J.~Sun, ``Deep residual learning for image recognition,'' in {\em Proceedings of the IEEE conference on computer vision and pattern recognition}, pp.~770--778, 2016.

\bibitem{10325570}
K.~Yamane, Y.~Saigusa, S.~Sakaino, and T.~Tsuji, ``Soft and rigid object grasping with cross-structure hand using bilateral control-based imitation learning,'' {\em IEEE Robotics and Automation Letters}, vol.~9, no.~2, pp.~1198--1205, 2024.

\bibitem{saigusa2022imitation}
Y.~Saigusa, S.~Sakaino, and T.~Tsuji, ``Imitation learning for nonprehensile manipulation through self-supervised learning considering motion speed,'' {\em IEEE Access}, vol.~10, pp.~68291--68306, 2022.

\bibitem{sakaino2011multi}
S.~Sakaino, T.~Sato, and K.~Ohnishi, ``Multi-dof micro-macro bilateral controller using oblique coordinate control,'' {\em IEEE Transactions on Industrial Informatics}, vol.~7, no.~3, pp.~446--454, 2011.

\bibitem{shikata2023modal}
K.~Shikata and S.~Katsura, ``Modal space control of bilateral system with elasticity for stable contact motion,'' {\em IEEJ Journal of Industry Applications}, vol.~12, no.~2, pp.~131--144, 2023.

\bibitem{kingma2014adam}
D.~P. Kingma and J.~Ba, ``Adam: A method for stochastic optimization,'' {\em Published as a conference paper at the 3rd International Conference for Learning Representations}, 2015.

\end{thebibliography}

\end{document}